%% file: samplepaper.tex
\documentclass[runningheads]{llncs}
\usepackage[T1]{fontenc}
\usepackage[shortlabels]{enumitem}
\usepackage{graphicx}
\usepackage{amssymb}
\usepackage[colorinlistoftodos,prependcaption,textsize=tiny]{todonotes}
\begin{document}
\title{Label Dropout: Improved Deep Learning Echocardiography Segmentation Using Multiple Datasets With Domain Shift and Partial Labelling}

\author{Iman Islam\inst{1} \and Esther Puyol-Ant\'{o}n\inst{1} \and Bram Ruijsink\inst{1}
Andrew J. Reader\inst{1} \and
Andrew P. King\inst{1}}
\authorrunning{I. Islam et al.}
\titlerunning{Label Dropout}

\institute{School of Biomedical Engineering {\&} Imaging Sciences, King's College London, UK}
%



\maketitle              
\begin{abstract}

\input{Sections/abstract}

\end{abstract}

\section{Introduction}
\input{Sections/introduction}

\section{Materials and Methods}
\label{sect:materials}
\input{Sections/materials}
\input{Sections/methods}


\section{Experiments and Results}
\label{sect:expts}
\input{Sections/experiments}

\section{Discussion and Conclusion}
\input{Sections/discussion}

\begin{credits}
\subsubsection{\ackname}
\input{Sections/acknowledgements}
\end{credits}
%
%
%
\bibliography{references}
\bibliographystyle{splncs04}

\end{document}

%% file: Sections/abstract.tex
Echocardiography (echo) is the first imaging modality used when assessing cardiac function. The measurement of functional biomarkers from echo relies upon the segmentation of cardiac structures and deep learning models have been proposed to automate the segmentation process. However, in order to translate these tools to widespread clinical use it is important that the segmentation models are robust to a wide variety of images (e.g. acquired from different scanners, by operators with different levels of expertise etc.). To achieve this level of robustness it is necessary that the models are trained with multiple diverse datasets. A significant challenge faced when training with multiple diverse datasets is the variation in label presence, i.e. the combined data are often \emph{partially-labelled}. Adaptations of the cross entropy loss function have been proposed to deal with partially labelled data. In this paper we show that training naively with such a loss function and multiple diverse datasets can lead to a form of shortcut learning, where the model associates label presence with domain characteristics, leading to a drop in performance. To address this problem, we propose  a novel \emph{label dropout} scheme to break the link between  domain characteristics and the presence or absence of labels. We demonstrate that label dropout improves echo segmentation Dice score by 62\% and 25\% on two cardiac structures when training using multiple diverse partially labelled datasets.

%% file: Sections/introduction.tex
Echocardiography  (echo) is the first imaging examination carried out when assessing cardiac function. Based on segmentations of the cardiac structures from echo images, useful biomarkers can be extracted to measure the function of the heart for diagnosis and treatment management. Deep learning models have been proposed to automate this segmentation process \cite{ghorbani_deep_2020,leclerc_deep_2019,ouyang_video-based_2020,painchaud_echocardiography_2022,puyol-anton_ai-enabled_2022,tromp_automated_2022}, but for clinical translation it is important that such models are robust to the wide variation in image characteristics that will be encountered in the real world (e.g. different scanners and operator levels of expertise etc.). In other applications, such as cardiac magnetic resonance, segmentation models have been trained using diverse data sources and shown to be robust to such variations \cite{mariscal-harana_artificial_2023}. In this work we aim to train a similarly robust model for echo segmentation.

One significant challenge that must be overcome when training with diverse datasets is the variation in label presence in the training data. For instance, a number of public datasets exist for training echo segmentation models but the manually defined labels present are different.  A summary of these datasets can be found in Table \ref{tab:datasum}. This means that the combination of these datasets will be \emph{partially labelled}, i.e. not all structures will be labelled in all training samples.

Training naively with partially labelled datasets such as these can cause a conflict in the supervision due to structures being labelled as foreground in some samples and background in others. A number of methods have been proposed to deal with such data by modifying the loss function to deal with the conflict. 
For example, Petit et al \cite{petit2018handling} proposed a loss function that summed the binary cross entropy (BCE) losses for each foreground class. For samples with missing ground truth labels, only the BCE terms with ground truth labels were computed.
A related approach was proposed by Shi et al \cite{shi_marginal_2021}. Their work proposed a marginal loss, which was a modification of the categorical cross entropy (CCE) loss function, in which any missing label was merged with the background class. In these merged regions, the loss function essentially considered both the missing label and the background to be a correct prediction.
Finally, Mariscal-Harana et al \cite{mariscal-harana_artificial_2023} proposed the adaptive loss, which was similar in concept to the approach of Petit et al \cite{petit2018handling},
but removed the loss for the label not present in the ground truth from the CCE loss calculation.

In this work, we show for the first time through a series of qualitative and quantitative experiments that when training an echo segmentation model using a loss function designed for use with partially labelled data, a form of shortcut learning can occur which leads the model to associate image characteristics with label presence. This causes a significant drop in performance for the missing labels. We propose a novel approach to address this problem,  called \emph{label dropout}, which aims to break the link between image characteristics and label presence and hence prevent the shortcut learning. We demonstrate that label dropout significantly improves segmentation performance when training using multiple diverse partially labelled echo datasets. 

%% file: Sections/materials.tex
\subsubsection{Datasets:} Three publicly available 2D echo datasets were exploited in this work (see Table \ref{tab:datasum}).
CAMUS and EchoNet Dynamic contain images at end diastole (ED) and end systole (ES) for each subject. Unity Imaging does not provide information at the subject level.
Upon manual review, several of the Unity Imaging segmentations were discarded due to the left ventricular myocardium (LVM) being overlabelled into the right ventricular myocardium. Therefore, we utilised 400 mostly apical 2-chamber images out of the 1504 available.
%
The extraneous information outside the echo cone in the Unity Imaging and EchoNet Dynamic images were removed before use in our experiments using an nnU-Net \cite{isensee_nnu-net_2021} model trained to segment the ultrasound cone. After pre-processing, all images were resized to $256\times256$.
\input{Tables/datasum}

%% file: Tables/datasum.tex
\begin{table}
    \centering
    \caption{Summary of the three datasets used in this work showing the number of subjects, number of images, ground truth segmentation labels present and image views. LV = left ventricle, LVM = left ventricular myocardium, LA = left atrium, A2C = apical 2-chamber, A4C = apical 4-chamber.}
    \begin{tabular}{lcccccc} \cline{1-7}
         &  No. of subjects &  No. of images &  LV&  LVM&  LA& Image view(s) \\ \cline{1-7}
         CAMUS \cite{leclerc_deep_2019} &  500&  1000&  \checkmark&  \checkmark&  \checkmark& A2C + A4C \\  
         Unity Imaging \cite{huang_fix--step_2023} &  - &  400&  \checkmark&  \checkmark&  \checkmark& A2C \\ 
         EchoNet Dynamic \cite{ouyang_video-based_2020} &  10024&  20048&  \checkmark&  &  & A2C + A4C \\ \hline
    \end{tabular}
    \label{tab:datasum}
\end{table}

%% file: Sections/methods.tex
\subsubsection{Baseline segmentation models:}
A U-Net \cite{ronneberger_u-net_2015} was used as the baseline segmentation model.
All models were trained using a stochastic gradient descent optimizer with a variable learning rate and a Nestorov momentum of 0.9 for 500 epochs. The initial learning rate and batch size was selected using a grid search and the model with the best foreground Dice score on the validation set was used for evaluation on the test set.
Models were trained using two different loss functions. First, \emph{standard loss} models were trained using a standard CCE loss function calculated over all classes including the background.
Second, we trained \emph{adaptive loss} models using the adaptive cross entropy loss proposed in \cite{mariscal-harana_artificial_2023}. This loss was implemented by removing the labels which are missing from the ground truth from the predicted segmentation, eliminating their contribution to the loss. 
Data augmentation was applied on the fly when training some models as specified in Section \ref{sect:expts}. The following augmentations were applied: scaling, rotation, Gaussian blur, brightness and contrast adjustment. The data were randomly split into 80\%/10\%/10\% for the training, validation and test sets.

\subsubsection{Label dropout:}
As we will show in Section \ref{sect:expts}, a significant drop in performance of the adaptive loss model was observed when training with partially labelled datasets which have a domain shift between them. It was hypothesised that this was due to the model learning to associate domain specific characteristics with the presence of labels. Therefore, we propose the label dropout scheme, which aims to break the link between domain characteristics and the presence or absence of certain labels. In label dropout, we introduce a probability of a label being removed (i.e. set to  background) from the ground truth mask of each sample during training. 
Each training sample is considered individually on the fly during training.
Label dropout is applied only to partially labelled classes (i.e. those that are missing in some ground truth masks).
In the case of there being multiple missing partially labelled classes, only one of them is dropped out, with an equal chance of each of them being chosen.

%% file: Sections/experiments.tex
This section will detail a series of experiments which aim to illustrate the problem when training with multiple diverse partially labelled datasets, as well as the effectiveness of our proposed label dropout scheme in overcoming this problem.
\subsubsection{Experiment 1 - The need for training with multiple diverse datasets:}

\begin{figure}[h]
    \centering
    \includegraphics[width=2.0in]{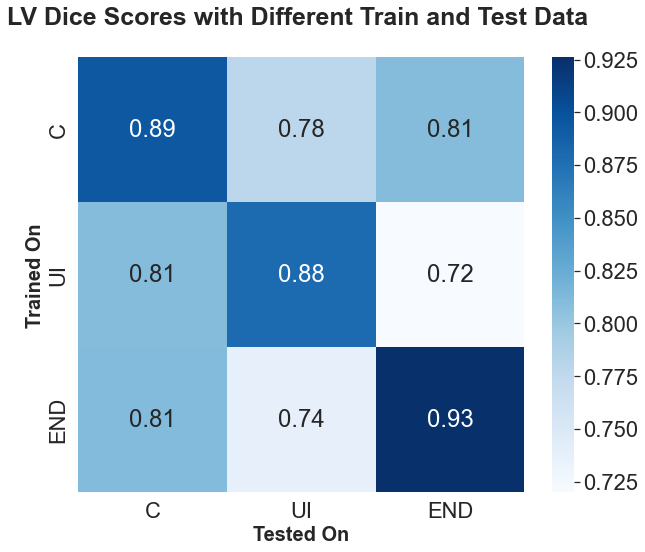}
    \caption{Experiment 1: Test Dice scores achieved by training and evaluating intra-domain and cross-domain LV segmentation models using three different datasets. C = CAMUS \cite{leclerc_deep_2019}, UI = Unity Imaging \cite{huang_fix--step_2023}, END = EchoNet Dynamic \cite{ouyang_video-based_2020}.}
    \label{fig:heatmap}
\end{figure}

We first illustrate the need to train echo segmentation models using multiple diverse datasets. In this experiment we trained left ventricle (LV) segmentation models with data augmentation using each of the datasets previously described in Section \ref{sect:materials}. Each model was evaluated on each of the three datasets.
Fig. \ref{fig:heatmap} shows the test set Dice scores achieved for each evaluation. As can be seen, the models perform worse when tested on datasets they were not trained on. Therefore, training models using multiple diverse datasets is necessary to improve the generalisability of echo segmentation models.

\subsubsection{Experiment 2 - Training using a combination of three diverse partially labelled echo datasets:}

The purpose of this experiment was to illustrate the problem of using the standard loss model when training with diverse partially labelled datasets and explore if using the adaptive loss model would lead to satisfactory results.
We again used all three datasets in this experiment but this time combined them into a single training dataset. Therefore, the training dataset was partially labelled and highly diverse including various cone shapes and sizes, intensity inhomogeneities within the cones and differing contrasts.

\begin{figure}[hbt!]
    \centering
    \includegraphics[width=4.2in]{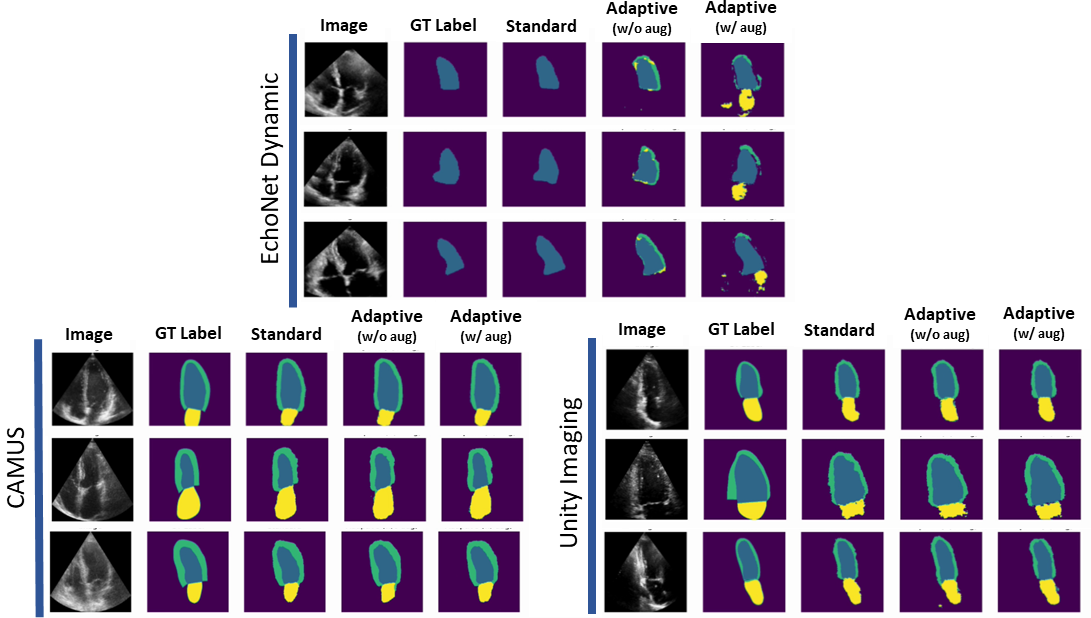}
    \caption{Experiment 2: Example test results from the three datasets. From left to right: image, ground truth segmentation and model predictions using standard loss model, adaptive loss without augmentation and adaptive loss with augmentation. }
    \label{fig:expt1res}
\end{figure}

Using these data, we trained three different models to segment the LV, LVM and LA: a standard loss model with augmentation and adaptive loss models with and without augmentation. Fig. \ref{fig:expt1res} shows a representative sample  of the results for this experiment. The difference in quality between the model predictions for the LV, where ground truth masks were always available is minimal.
However, the standard loss model completely fails to predict the LVM and LA in the EchoNet Dynamic dataset whilst predicting all three labels in the other datasets. This shows a form of shortcut learning in which the model has learnt to associate domain characteristics with label presence or absence, resulting in a model that only predicts labels which were present in the datasets' ground truths. Furthermore, contrary to expectations, the adaptive loss models do not produce accurate predictions for structures with missing labels. Even with augmentation, the predictions are less than satisfactory, particularly for the EchoNet Dynamic dataset.
Data augmentation improved the results to some extent by providing a way to reduce the impacts of the domain shifts, however it did not completely overcome the impact of the shortcut learning.

\subsubsection{Experiment 3 - Investigating the adaptive loss in a controlled experiment:}

The purpose of this experiment was to further investigate our hypothesis that domain shift has led to shortcut learning in the adaptive loss model.
Here, we investigate the viability of the adaptive loss model in a controlled environment, with no domain shift between the differently labelled samples. To achieve this, ground truth labels were artificially removed from a subset of samples from the CAMUS dataset. Three models were trained and evaluated. The first was a benchmark model viewed as the best achievable performance with no partial labelling, using a standard loss. Then, a standard loss model and an adaptive loss model (both without augmentation) were trained with the LVM label artificially removed from 50\% of the training data. 

\begin{figure}[hbt!]
    \centering
    \includegraphics[width=4.5in]{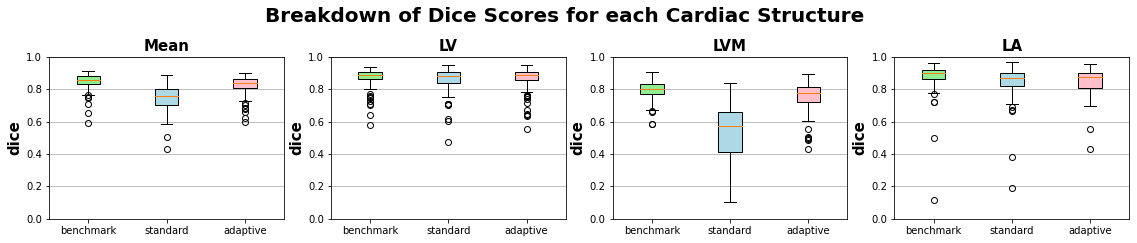}
    \caption{Experiment 3: Test set results when training using only the CAMUS dataset with 50\% of LVM labels removed. Box plots show Dice coefficients for each segmented structure and the overall mean. Green = benchmark, blue = standard, pink = adaptive.}
    \label{fig:expt2res}
\end{figure}

The test set results are shown in Fig. \ref{fig:expt2res}. The benchmark model, standard loss model and adaptive loss model achieved mean foreground Dice scores of $0.873\pm0.05$, $0.803\pm0.06$, and $0.863\pm0.05$ respectively. Therefore, in this controlled experiment, although the adaptive loss model was trained with missing labels, it performed comparably to the benchmark model. The standard loss model suffered a significant drop in performance due to the conflict in supervision information. The box plots in Fig. \ref{fig:expt2res} show the breakdown of the Dice scores for each class, with the adaptive loss model achieving a significantly higher Dice score for the LVM compared to the standard loss model. 

This experiment shows the viability of the adaptive loss as a solution to the problem of partial labelling. It also supports our hypothesis that the lack of improvement of the adaptive loss model in Experiment 2 was due to the domain shift between the datasets leading to a form of shortcut learning. 

\subsubsection{Experiment 4 - Label dropout:}

Experiment 3 showed that the adaptive loss  has the potential to deal with partially labelled training data, but it did not produce clinically acceptable segmentations in Experiment 2 when there were domain shifts between the differently labelled datasets. In this experiment, the utility of our proposed label dropout scheme in addressing this problem is explored.
Two datasets were used in the first part of this experiment: CAMUS and Unity Imaging. 
The LA was artificially removed in all Unity Imaging ground truth masks in the training set to produce a combined training set with partial labels and a domain shift.
Two types of model were trained: adaptive loss models with label dropout using the partially labelled data, and a benchmark model with a standard loss and fully labelled data, which represents the best achievable performance. For the label dropout model, different probabilities for label dropout ranging from 0.0 to 1.0 in steps of 0.1 were tested. This is the probability of the LA being dropped out from the ground-truth segmentation for each training pair during training. Augmentation was used for training all models in this experiment.

\begin{figure}[hbt!]
    \centering
    \includegraphics[width=4.5in]{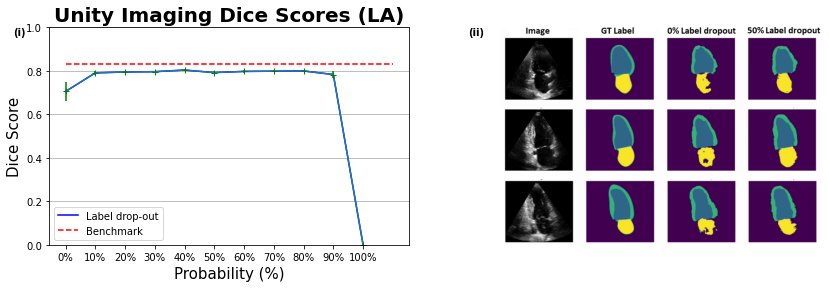}
    \caption{Experiment 4: Label Dropout. (i) Test set Dice scores from models trained with different probabilities of label dropout on the LA for images from the Unity Imaging dataset. Models were trained three times with different random seeds and the error bars show the mean and standard deviation of the results. Benchmark was trained with a standard loss using fully labelled data. (ii) Sample results  on the Unity Imaging dataset. From left to right: image, ground truth segmentation and model predictions without label dropout and with 50\% label dropout.}
    \label{fig:expt3res}
\end{figure}

Fig. \ref{fig:expt3res} shows the test set results of this experiment. The plot clearly shows the improved performance when using the label dropout technique which persists across a range of dropout probabilities.
Note that the 0\% label dropout model is equivalent to the adaptive loss model without label dropout. 
The benchmark model and the 0\% label dropout models achieved Dice scores of 0.83 and 0.71 respectively. The Dice scores of the adaptive loss models with label dropout were approximately 0.8. 100\% label dropout means that the LA is never seen during training.
Fig. \ref{fig:expt3res} shows how introducing label dropout, even with a very low probability of dropout at 10\%, helps to break the link between domain characteristics and the presence of labels.
In Fig. \ref{fig:expt3res}, sample model predictions from this experiment are  also displayed. When there is no label dropout, the LA in the Unity Imaging dataset has a visibly worse segmentation compared to the adaptive loss model with 50\% label dropout.
This experiment shows that label dropout can improve model performance when training with diverse partially labelled datasets and a Wilcoxon Signed Rank Test confirms this with a p-value of <0.01.
The Dice scores for the LV and LVM were similar for both models for each dataset.

\begin{figure}[hbt!]
    \centering
    \includegraphics[width=4.5in]{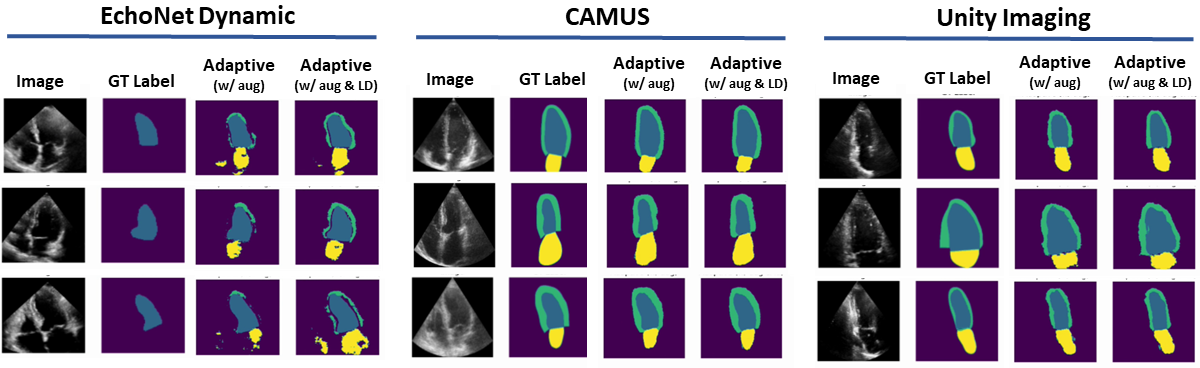}
    \caption{Repetition of Experiment 2 with label dropout. Randomly selected test set results when training with all 3 datasets using label dropout (LD). From left to right: image, ground truth segmentation and model predictions using adaptive loss with augmentation and adaptive loss with augmentation and label dropout.}
    \label{fig:finalresults}
\end{figure}

As a final experiment, we repeated Experiment 2 but now using the combination of all three datasets using label dropout. Sample test set results are shown in Fig. \ref{fig:finalresults}.  We see a visible improvement in the performance of the adaptive loss model when using label dropout, further supporting our hypothesis that a form of shortcut learning can negatively affect the performance of the adaptive loss when training using diverse partially labelled datasets.

To quantify these improvements, the LVM and LA were manually segmented by a trained observer (and checked by a cardiologist) in a random sample of 20 images from EchoNet Dynamic test set and some key results are shown in Table \ref{tab:finalresults}. The LVM and LA Dice scores represent Dice increases of 0.199 and 0.136 (or 62\% and 25\%) respectively, supported by a Wilcoxon Signed Rank Test with a p-value of <0.05. Model predictions for 10 random images from these 20 are displayed in Supplementary Fig. 1.

\begin{table}[hbt]
    \caption{Mean test Dice scores over 20 random images from EchoNet Dynamic when trained on EchoNet Dynamic, CAMUS and Unity Imaging data. Dice scores are displayed for each of the three cardiac structures segmented by the adaptive loss model with and without label dropout. Aug = augmentation, LD = label dropout, LV = left ventricle, LVM = left ventricular myocardium, LA = left atrium.}
    \begin{tabular*}{\textwidth}{@{\extracolsep\fill}|l|c|c|c|} \hline
         &  LV&  LVM& LA\\ \hline  
         Adaptive (w/ aug)&  0.892&  0.319& 0.553\\ 
         Adaptive (w/ aug \& LD)&  0.910&  0.518& 0.689\\  \hline
    \end{tabular*}
    
    \label{tab:finalresults}
\end{table}

%% file: Sections/discussion.tex
This paper has made two significant contributions: (i) we have highlighted for the first time that state-of-the-art approaches for dealing with partially labelled segmentation data can be negatively affected by a form of shortcut learning when trained with datasets featuring domain shift, (ii) we have proposed a new label dropout technique for dealing with this problem.
For contribution (i), we note that the adaptive loss that we employed to deal with the partially labelled data was shown to work effectively in an experimental environment with no domain shift (Experiment 3) and has previously been shown to be effective when there was domain shift but no relationship between domain characteristics and label presence (e.g. occasionally missing LVM at ES in cine cardiac magnetic resonance \cite{mariscal-harana_artificial_2023}). Thus, we conclude that its poorer performance in Experiment 2 was due to the presence of such a relationship.
For contribution (ii), label dropout was shown to improve model performance in Experiment 4. It is noticeable that the Dice score for the label dropout scheme plateaus when the label dropout is introduced. We speculate that this could be because, after a certain number of epochs, the model eventually sees all images with all labels. 

We believe that this work is important for training robust segmentation models.
When combining multiple diverse echocardiography segmentation datasets, the resulting training datasets are typically partially labelled and therefore this technique could
allow the training of more generalisable models.

Further work will include investigating the impact of label dropout on different network architectures, such as transformers \cite{cao_swin-unet_2021}, as well as alternative strategies for dealing with partial labels, such as the marginal loss \cite{shi_marginal_2021}. Furthermore, the label dropout technique will also be evaluated on datasets from different imaging modalities.

%% file: Sections/acknowledgements.tex
We would like to acknowledge funding from the EPSRC Centre for Doctoral Training in Medical Imaging (EP/L015226/1).
